\documentclass[sigconf]{acmart}
\usepackage{makecell}
\usepackage{graphicx}
\usepackage{subcaption}
\usepackage{booktabs}
\usepackage[inline]{enumitem}
\usepackage{multirow}
\usepackage{xcolor}
\usepackage{multirow}
\usepackage{url}
\usepackage{bbm}
\usepackage{textcomp}
\usepackage[titlenumbered,ruled,noend]{algorithm2e}
\usepackage{mathtools}
\usepackage{bbold}
\usepackage{pgfplots}
\usepackage{tikz}

\usepackage{amsthm}
\usepackage{array}
\newtheorem*{definition}{Definition}
\newtheorem*{problem}{Problem Description}
\newcolumntype{x}[1]{>{\centering\arraybackslash\hspace{0pt}}p{#1}}

\AtBeginDocument{%
  \providecommand\BibTeX{{%
    \normalfont B\kern-0.5em{\scshape i\kern-0.25em b}\kern-0.8em\TeX}}}

\setcopyright{acmcopyright}
\acmConference[KDD '22]{Proceedings of the 28th ACM SIGKDD Conference on Knowledge Discovery and Data Mining}{August 14--18, 2022}{Washington DC, DC, USA}
\acmBooktitle{Proceedings of the 28th ACM SIGKDD Conference on Knowledge Discovery and Data Mining (KDD '22), August 14--18, 2022, Washington DC, DC, USA
}
\copyrightyear{2022}
\acmYear{2022}
\acmDOI{10.1145/3534678.3539344}

\begin{document}

\title{Framing Algorithmic Recourse for Anomaly Detection}
\author{Debanjan Datta}
\email{ddatta@vt.edu}
\affiliation{
   \institution{Virginia Tech}
   \city{Arlington}
   \state{Virginia}
   \country{United States}
}


\author{Feng Chen}
\email{feng.chen@utdallas.edu}
\affiliation{
  \institution{University of Texas, Dallas}
  \city{Dallas}
\state{Texas}
\country{United States}
}
\author{Naren Ramakrishnan}
\email{naren@vt.edu}
\affiliation{
   \institution{Virginia Tech}
   \city{Arlington}
   \state{Virginia}
   \country{United States}
}

\renewcommand{\shortauthors}{Datta, et al.}

\begin{abstract}
The problem of algorithmic recourse has been explored for supervised machine learning models, to provide more interpretable, transparent and robust outcomes from decision support systems.
An unexplored area is that of algorithmic recourse for anomaly detection, specifically for tabular data with only discrete feature values. Here the problem is to present a set of counterfactuals that are deemed normal by the underlying anomaly detection model so that applications can utilize this information for explanation purposes or to recommend countermeasures.
We present an approach---\textbf{C}ontext preserving \textbf{A}lgorithmic \textbf{R}ecourse for \textbf{A}nomalies in \textbf{T}abular data (\textit{CARAT}), that is effective, scalable, and  agnostic to the underlying anomaly detection model.
\textit{CARAT} uses a transformer based encoder-decoder model to explain an anomaly by finding features with low likelihood.
Subsequently semantically coherent counterfactuals are generated by modifying the highlighted features, using the overall context of features in the anomalous instance(s). Extensive experiments help demonstrate the efficacy
of \textit{CARAT}.
\end{abstract}




\maketitle
\section{Introduction}
Algorithmic recourse can be defined as a a set of actions or changes that can change the outcome for a data instance with respect to a machine learning model, typically from an unfavorable outcome to a favorable one~\cite{joshi2019towards}.
This is an important and challenging task with practical applicability in domains such as healthcare, hiring, insurance, and commerce that incorporate machine learning models into decision support systems~\cite{prosperi2020causal,karimi2020survey}.
Algorithmic recourse is closely related to explainability, specifically counterfactual explanations that are important to improve fairness, transparency, and trust in output of machine learning (ML) models. 
Indeed the most cited and intuitive explanation of algorithmic recourse presents an example how to change input features of bank loan application decided by a black-box ML algorithm to obtain a favorable outcome~\cite{karimi2021algorithmic}.

Although the primary focus of algorithmic recourse has been in supervised learning contexts~\cite{mothilal2020explaining}, specifically classification based scenarios, it is also applicable in other scenarios. 
In this work, we address the research of how to frame algorithmic recourse for outcomes of unsupervised anomaly detection.
Specifically, we seek to obtain a set of actions to modify the feature values of a data instance deemed anomalous by a black-box anomaly detection model such that it is no longer anomalous.
A motivating example would be the case of a shipment transaction that is flagged as suspicious or illegal by a monitoring system employing anomaly detection, and our exploring what needs to be modified in this transaction to no longer merit that outcome.
An entity such as a trading company might seek to address its future shipment patterns, by adjusting routes, products or suppliers to avoid getting flagged as potentially fraudulent -- thus motivating the problem of algorithmic recourse for anomaly detection.

Algorithmic recourse for anomaly detection has some factors that differentiates it from the classification based scenario due to the underlying ML model w.r.t. which one tries to achieve a different but favorable outcome.
While classification models are supervised, anomaly detection models are mostly unsupervised, and archetypes of anomalies are difficult to determine and are application scenario dependant. 
Prior works consider tabular data for algorithmic recourse in the context of classification~\cite{karimi2020survey,rawal2020beyond}, and use comparatively simpler datasets where features are mostly real-valued. 
We explore the scenario where features are strictly categorical with high dimensionality (cardinality), such as found in real world data from commerce, communication and shipping~\cite{cao2018collective,datta2020detecting}.
Concepts such as \textit{proximity} in the context of counterfactuals are simpler to define for real-valued data.
Moreover, metrics used in classification specific algorithmic recourse do not directly translate to the scenario of anomaly detection.
Our key contributions in this work are:
\begin{enumerate}[label=(\textbf{\roman*})]
\item A novel formulation for the unexplored problem of algorithmic recourse for unsupervised anomaly detection.
\item A novel approach \textit{CARAT} to generate counterfactuals for anomalies in tabular data with categorical features. 
\textit{CARAT} is demonstrated to be effective, scalable and agnostic to the underlying anomaly detection model.
\item A new set of metrics that can effectively quantify the quality of the generated counterfactuals w.r.t. multiple objectives.
\item Empirical results on multiple real world shipment datasets along with a case study highlighting the practical utility of our approach.
\end{enumerate}


\section{Preliminaries}\label{sec:preliminaries}
Tabular data with strictly categorical attributes can be formally represented in terms of \textit{domains} and \textit{entities}~\cite{datta2020detecting}. 
A \textit{domain} or attribute or categorical feature is defined as a set of elements sharing a common property, e.g. \textit{Port}. 
A domain consists of a set of \textit{entities} which are the set of possible values for the categorical variable, e.g. \textit{Port}: \{ Baltimore, New York, \dots\}.
\textit{Context}~\cite{datta2020detecting} is defined as the reference group of entities with which an entity occurs, implying an entity can be present in multiple contexts.
A data instance (record) is anomalous if it contains unexpected co-occurrence among two or more of its entities~\cite{das2007detecting,hu2016embedding,datta2020detecting}.

\begin{definition}[Anomalous Record]
An anomalous record is a record where certain domains have entity values that are not consistent with the remaining entity values, termed as the context, with respect to the expected data distribution. 
\end{definition}

\textit{Explanation} for a model typically refers to an attempt to convey the internal state or logic of an algorithm that leads to a decision~\cite{wachter2017counterfactual}.
Closely related to the idea of explanations are counterfactuals.
\textit{Counterfactuals} are hypothetical examples that demonstrate to an user how a different and desired prediction can be obtained.
\textit{Algorithmic recourse} has been defined as an actionable set of changes that can be made to change the prediction of a system with respect to a data instance from an unfavourable one to a desirable one~\cite{joshi2019towards}.
The idea is to change one or more of the feature values of the input in an feasible manner in order to produce a favorable outcome. 
Algorithmic recourse has been explored in the context of mostly classification problems, with a generalized binary outcome scenario.
Algorithmic recourse for anomaly detection is an important yet mostly unexplored problem.
In this work, the hypothetical instances that are the result of algorithmic recourse on a data instance are referred to as \textit{counterfactuals} or \textit{recourse candidates}. 
It is important to note that while \textit{counterfactual explanations} provide explanations through contrasting examples, \textit{algorithmic recourse} refers to the set of actions that provides the desired outcome.

While nominal points are assumed to be generated from an underlying data distribution $\mathcal{D}$, anomalies can be assumed to be generated from a different distribution $\mathcal{D'}$.
It can be hypothesized that an anomaly $\textbf{x}_a \sim \mathcal{D'}$, is generated from some $\textbf{x}_n \sim \mathcal{D}$ through some transformation function set $\mathcal{F}$, such that $\textbf{x}_a = \mathcal{F}(\textbf{x}_n) \sim \mathcal{D'}$. 
A simplifying view of $\mathcal{F}$ can be a process of feature value perturbation or corruption.
Therefore, we can also hypothesize that there exists some arbitrary function set $\mathcal{G}$, such that $\mathcal{G}(\textbf{x}_a) \sim \mathcal{D}$ and possibly $\mathcal{G}(\textbf{x}_a)\neq \textbf{x}_n$ --- which is emulated through algorithmic recourse. 

\begin{definition}[Algorithmic Recourse for Anomaly Detection]
Algorithmic recourse for anomaly detection can be defined as a set of actionable changes on an anomalous data instance, such that it is no longer considered an anomaly with respect to the underlying anomaly detection model. 
\end{definition}

Specifically, we consider the research question that given a row of tabular data, with strictly categorical values, which is deemed anomalous by an anomaly detection model $\mathcal{M}_{AD}$ --- how can we generate a set of hypothetical records such which would be deemed normal by $\mathcal{M}_{AD}$. 
In this setting, without loss of generality we consider $\mathcal{M}_{AD}$ to be 
\begin{enumerate*}[label=(\roman*)]
\item trained using a training set which is assumed to be clean~\cite{chen2016entity,datta2020detecting},
\item a likelihood based model that produces real-valued scores,
\item a queriable black box model
\end{enumerate*}
\begin{problem}[]
Given data instance $\textbf{x}_a$ which is deemed anomalous by a given \textit{anomaly detection} model $\mathcal{M}_{AD}$, the objective is to generate a set of counterfactuals $Y_{cf}$ such that $\textbf{x}_{cf} \in Y_{cf}$ is not an anomaly according to $\mathcal{M}_{AD}$.
\end{problem}

Since in the case of unsupervised anomaly detection an application or dataset specific threshold is often used which is difficult to determine, we can relax the definition of recourse to a obtain a set of counterfactuals $Y$ such that $\textbf{x}_{cf}\in Y_{cf}$ are ranked lower by $\mathcal{M}_{AD}$ in terms of anomaly score. 
There are multiple objectives that require optimization to obtain counterfactuals that satisfy different criterion~\cite{karimi2020model} such as \textit{sparsity}, \textit{diversity}, \textit{prolixity}~\cite{keane2020good}, \textit{proximity} to the anomalous record, low \textit{cost} to the end user, along with \textit{feasibility}, \textit{actionability} and \textit{non-discriminatory} nature which depend on application scenario. 
Since we address a general scenario without apriori application specific knowledge, some of these problem specific objectives such as user specific cost or feasibility are not applicable.
We consider the key criterion such as validity, diversity and sparsity and discuss them on evaluation metrics in Section~\ref{sec:metrics}.

\begin{figure*}[t]
\begin{subfigure}[b]{0.49\textwidth}
  \centering
  \includegraphics[width=0.975\linewidth]{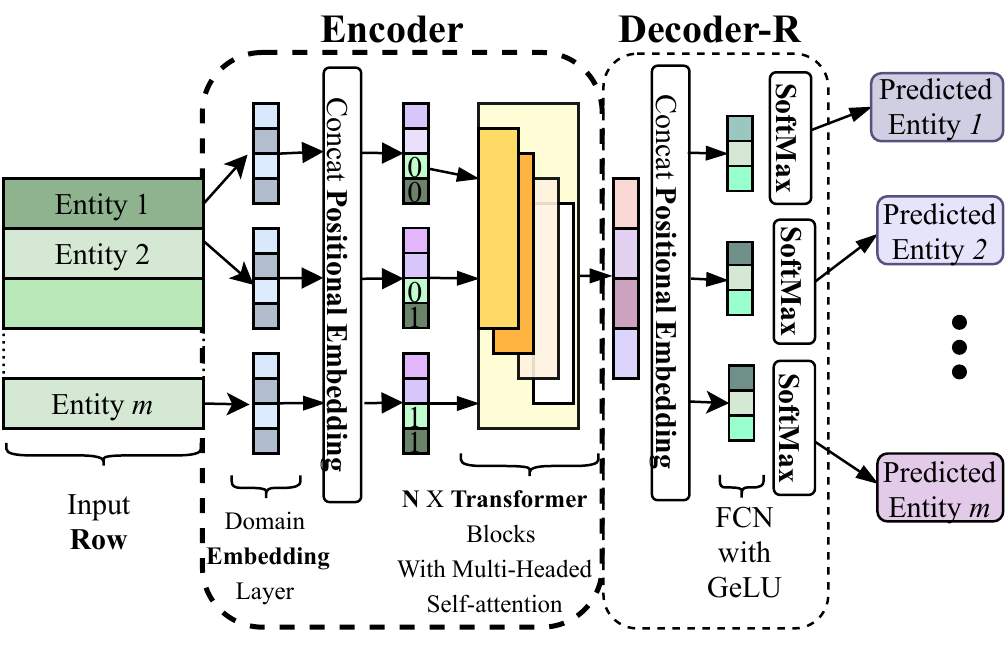}
  \caption{Architecture of the Encoder and the Decoder-R used in the first phase of pretraining the Encoder to learn the representation the entities and capture overall context of the record (Section~\ref{sec:pretrain}). }
  \label{Lill fig:sub-first}
\end{subfigure}\hfill
\begin{subfigure}[b]{0.5\textwidth}
  \centering
  \includegraphics[ width=0.975\linewidth]{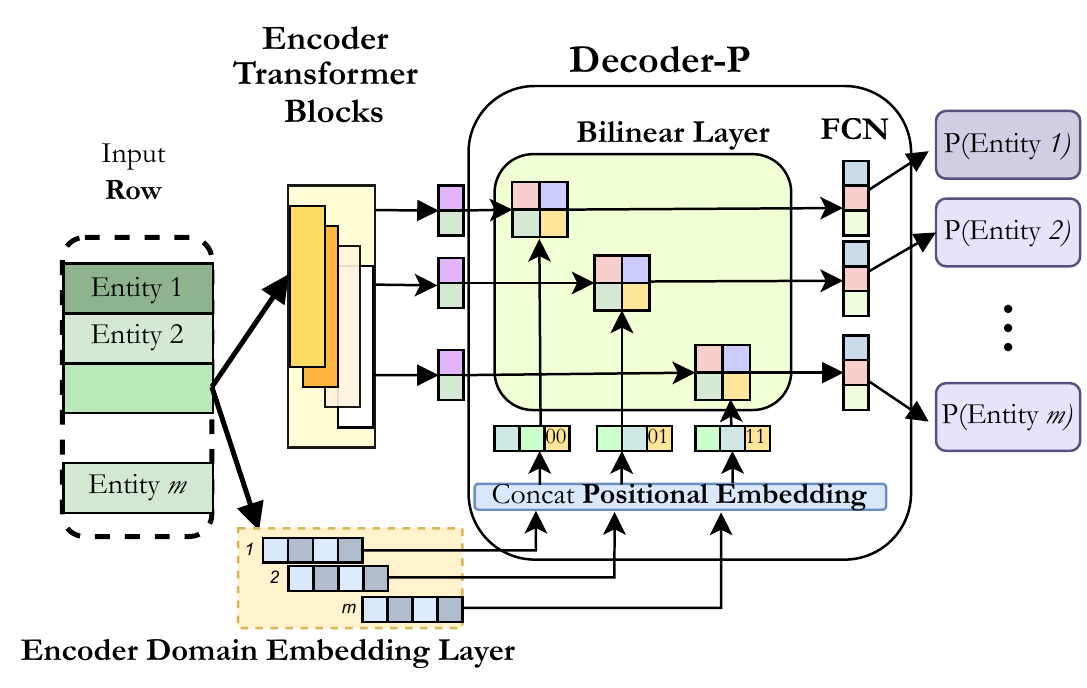}
  \caption{Architecture of Decoder-P, which takes both the embeddings and the contextual representation  of entities from the  encoder, and predicts likelihood of each entity in the record (Section~\ref{sec:decoder-p}) }
  \label{Lill fig:sub-second}
\end{subfigure}\hfill

\vspace{-3mm}
\caption{Architecture of the Explainer model in \textit{CARAT}, comprising of the encoder and decoder-P, which captures likelihood of each entity in the input record.}
\label{fig:xformer}
\vspace{-3mm}
\end{figure*}
\vspace{-2mm}
\section{Related Work}
In this section, we discuss the prior literature that explores the concepts of algorithmic recourse, counterfactuals and anomaly explanation, which are relevant in this discourse. 

Explainability in machine learning models has gained burgeoning research focus over the last decade due to the need for building trust and achieving transparency in decision support systems that often employ black-box models.
Post-hoc explanations through feature importance has been proposed to explain prediction of classifiers.
\textit{LIME}~\cite{ribeiro2016should} presents an approach to obtaining explanations through locally approximating a model  in the neighborhood of of a prediction of a prediction. 
~\textit{DeepLIFT}~\cite{shrikumar2017learning} proposed a method to decompose the prediction of a neural network by  recursively calculating the contributions by individual neurons. 
~\textit{SHAP}~\cite{SHAP}present a unified framework that assigns each feature an additive importance measure for a particular prediction, based on \textit{Shapley} Values.
\textit{InterpretM}L~\cite{nori2019interpretml} presents an unified framework for Ml interpretability.

There has been recent work on explaining outcome for anomaly detection models~\cite{antwarg2021explaining, macha2018explaining, yepmo2022anomaly}.
\textit{ACE}~\cite{zhang2019ace} proposes an approach for explaining anomalies in cybersecurity datasets.
While some some anomaly detection methods such as \textit{LODI}~\cite{dang2013local} and \textit{LOGP}~\cite{dang2014discriminative} provide feature importance to explain anomalies by design, most methods employ a post-hoc explanation approach.
DIFFI~\cite{carletti2020interpretable} provides explanations for outputs from an Isolation Forest.
Explanation through gradient based approaches have been proposed in anomaly detection based methods on neural networks~\cite{amarasinghe2018toward,nguyen2019gee,kauffmann2020towards}.

\citeauthor{karimi2020survey}~\cite{karimi2020survey} presents a comprehensive survey on algorithmic recourse.
\citeauthor{ustun2019actionable}~\cite{ustun2019actionable} introduced the notion of actionable recourse, that ensures that the counterfactuals are  obtained through appropriate feature value modification.
\textit{DiCE}~\cite{mothilal2020explaining} presents an a framework for generating and evaluating a diverse set of counterfactuals based on determinantal point processes. 
Neural network model based approaches for generating counterfactuals have also been proposed ~\cite{pawelczyk2020learning,mahajan2019preserving, chapman2021fimap}.
Causal reasoning has also been explored towards algorithmic recourse~\cite{karimi2020model,prosperi2020causal,karimi2021algorithmic,crupi2021counterfactual}.
Approaches for recourse based on heuristics, specifically using genetic algorithms have also been explored~\cite{sharma2019certifai, barredo2020plausible, dandl2020multi}.
To our knowledge, only one method RCEAA~\cite{RCEAA} has been proposed towards recourse in anomaly detection based on autoencoders with real valued inputs.


\section{Algorithmic Recourse Through Modeling Context}\label{sec:model}
Algorithmic recourse consists of two steps:
\begin{enumerate*}[label=(\roman*)]
    \item Understanding what is causing a data instance to be an anomaly,
    \item How to define a set of actions to modify the feature values in order to remedy the unfavorable outcome.
\end{enumerate*}
We propose \textbf{CARAT}: \textbf{C}ontext preserving \textbf{A}lgorithmic \textbf{R}ecourse for \textbf{A}nomalies in \textbf{T}abular Data that decomposes the task into these two sequential logical steps and address them.
\textit{CARAT} comprises of a model based approach to identify the presence of entities that causes the record to be anomalous,
and an algorithm to modify those feature values in the record for recourse.

\vspace{-2mm}

\subsection{Explainer Model}\label{sec:xformer}
Given a record or data instance with categorical features, the tuple of entities is anomalous when one or more of the entities are out-of-context with respect to the remaining entities~\cite{das2007detecting} with unexpected co-occurrence patterns.
We use a Transformer~\cite{vaswani2017attention} based architecture to jointly model the context of the entities of records. 
Transformers have been extensively utilized in other applications on text, image and tabular data~\cite{huang2020tabtransformer}. 
A record can be considered as a sequence of entities, without a predefined ordering of domains or any semantic interpretation of the relative ordering.
Transformer based architectures are appropriate for tabular data with categorical features since
\begin{enumerate*}[label=(\roman*)]
\item they can handle large cardinality values for each category and are scalable
\item can provide contextual representations of entities
\item can model context with a prespecified ordering of domains
\end{enumerate*} and do not consider any relative ordering among the domains (categories).
We adopt an encoder-decoder architecture, similar to language models~\cite{devlin-etal-2019-bert} with the objective to predict the likelihood of each entity in a given record with possible corruptions.
The predicted likelihood for each entity is conditioned on the context---implicitly capturing the pair-wise and higher order co-occurrence patterns among entities.

\vspace{-1mm}
\subsubsection{Pretrained Row Encoder}\label{sec:pretrain}
The encoder has a transformer based architecture and consists of multiple layers.
Sequential architectures are not effective for rows of tabular data where the relative ordering of entities (and domains) do not have any semantic interpretation.
To handle domains with large number of entities, a $m$ parallel domain specific embedding layers are used with same dimensionality.
To inform the model which domain an entity embedding vector belongs to, we utilize \textit{positional encoding}~\cite{devlin-etal-2019-bert} vectors which is concatenated to each of the entity embedding vectors.
The tuple of vectors is then passed to the subsequent transformer block comprising of multiple layers of transformers.

To train the encoder such that it learns contextual representation for each entity in a record, we require a corresponding decoder and training objective which we design as follows.
We refer to this decoder as \textit{decoder-R}, which is used for pretraining the encoder and not in the final objective.
\textit{Decoder-R} comprises of multiple fully connected layers and is trained to reconstruct data. 
In order to aid the network to retain information and reconstruct it accurately, the contextual entity embeddings from the encoder layer are augmented with positional vectors through concatenation, after the first fully connected layer.
Note that both the encoder and \textit{decoder-R} utilize positional encoding to indicate domain and help the model reconstruct the entity for a given domain using the contextual embedding. 
The remainder of the \textit{decoder-R} consists of $m$ parallel dense layers with GELU activation, with the last layer being \textit{softmax} to obtain the index of the entity for a specific domain.
Note that while the first transformation layer is shared for entity embedding of all domains, the latter layers are domain specific.
The encoder and \textit{decoder-R} are jointly trained, using a reconstruction based objective, similar to Masked Language Model where we randomly perturb or remove entities from records and train the model to predict the correct one from the partial context. 
The trained encoder captures the shallow embedding in the first layer as well as the contextual representation of entities in the record. 

\vspace{-1mm}
\subsubsection{Entity Likelihood Prediction Model}\label{sec:decoder-p}
The \textit{decoder-P} is designed to predict the likelihood of each entity in a record as output---using the outputs from the pretrained encoder as it's input.
The input to \textit{decoder-P} consists of 
\begin{enumerate*}[label=(\roman*)]
\item The embedding representation for the $j^{th}$ entity $e^j$, obtained from the first embedding layer of the encoder ($x_0^j \in R^d$) 
\item The contextual representation of the entity $e^j$, obtained from the last layer of the encoder, $z^j$. 
\end{enumerate*}
Domain specific \textit{positional encoding} vector ($p^j \in R^d $) is concatenated with $x_0^j$ to obtain $x^j \in R^{2d}$.
We want to capture the semantic coherence and interaction between $x^j$, and the contextual representation of the entity $z^j$.
To accomplish this we use a \textit{Bilinear} layer.
The output of this Bilinear layer is fed to a dense network with multiple hidden layers, and finally a sigmoid activation function to obtain a likelihood of whether an entity should occur in the given record. 
We utilize a simple 2-layered architecture with ReLU activation for this domain specific dense layer. 
Binary cross-entropy loss is used to train \textit{decoder-P}, keeping the weights of the pretrained encoder fixed.
The training of \textit{decoder-R} differs from \textit{decoder-P} due to the divergent objective.
We generate labelled samples from the training data, where we perturb samples with a probability $1-\alpha$. For $\alpha$ fraction of samples, the model is given unchanged records from the training set to enable it to recognize expected patterns and predict higher likelihood scores for co-occurring entities.
For the remaining samples, we randomly perturb one or more of its entities and task the decoder to recognize which of the entities have been perturbed.
It is important to note that the objective of the \textit{explainer} model is not anomaly detection, but  to predict the likelihoods of individual entities in a record.

\begin{figure}[tp]
    \centering
    \includegraphics[width=0.700\columnwidth]{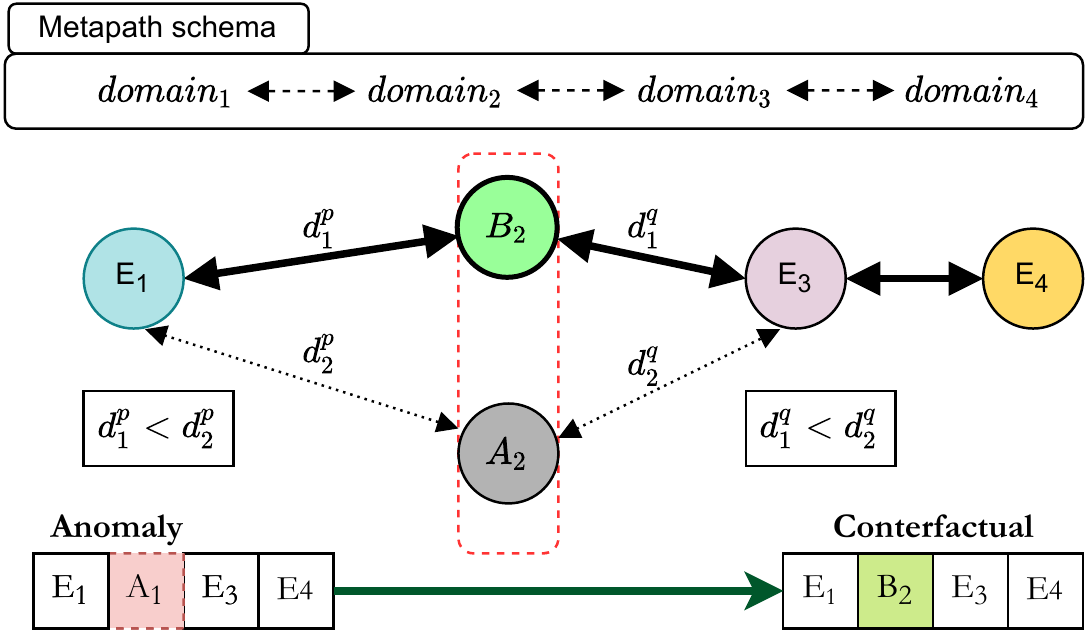}
    \caption{Finding counterfactuals through replacing a lower likelihood entity in a record with another entity based on semantic similarity with respect to the other entities. }
    \label{fig:semantic}
\end{figure}

\subsection{Generating Counterfactuals}

\begin{algorithm}[ht!]
\SetKwComment{Comment}{//}{}
\SetKwInOut{Input}{Input}
\SetKwInOut{Output}{Output}

\caption{Counterfactual generation in \textit{CARAT}}\label{alg:recourse_gen_xformer}
\Input{Anomalous record $\textbf{x}_a$; Explainer Model $\mathcal{M}_{t}$, Anomaly Detection model $\mathcal{M}_{AD}$, Set of metapaths $\mathrm{MP}:\{mp_1, mp_2..mp_q\}$,  K: No. of counterfactuals} 
\Output{Set of conterfactuals CF for $\textbf{x}_a$}

$\{d_{mod}\}$ $\gets$ Set of domains $d_j$ where $\mathcal{M}_{t}$: $P(e_j^a \in \textbf{x}_a) < 0.5$ \\
If $\{d_{mod}\} = \phi$ : $\{d_{mod}\}$ $\gets d_j$ where $min_j(P(e_j^a \in \textbf{x}_a))$  \Comment*[r]{Ensure at least a single domain is modified} 
\For {$d_i \in \{d_{mod}\}$ }{
    $C_i = \{\} $\Comment*[r]{Candidate entities to replace $e_i^a$} 
    \For{$mp_j \in \mathrm{MP}$}{
        $Q = \{\}$ \\
        \If{$d_i \in mp_j$ }{
            $S \gets Nbr(mp_j,d_i)$, such that $S\cap \{d_{mod}\} = \phi$ \\
            $Q \gets Q \cup S$ 
        }
        \For {each $d_q \in Q$}{
        $P \gets K\, Nearest\, Neighbors$ for $e_q^a \in d_i$ 
        \Comment*[r]{ Find entities for $d_i$ similar to other entities in $\textbf{x}_a$  }
        $C_i = C_i \cup P$ \Comment*[r]{Update candidate entities}
        }
    }
}
$CF \gets \{\}$ \\
\For{combinations of ($C_i,d_i \in \{d_{mod}\}$) for $\forall d_i \in\{d_{mod}\}$}{
    $\textbf{x}_{cf} \gets$ Replace $e_i^a \in d_i$ in $\textbf{x}_a$ with $c_i \in C_i$ \;
    $CF$ $\gets$ $CF$ $\cup$ $\textbf{x}_{cf}$
}
Score $\textbf{x}_{cf} \in CF$ using $\mathcal{M}_{AD}$ \;
$CF$ $\gets$ Least K anomalous records in $CF$ \;

\end{algorithm}

Records in tabular data with categorical features can be considered as tuple of entities, with data specific inherent relationships between the domains (attributes).
For instance in the case of shipment records, products being shipped are closely related to the company trading them and their origin.
Many real-life applications involve tabular data which can be represented as a heterogeneous graph or Heterogeneous Information Network (HIN). 
A HIN~\cite{sun2011pathsim} is formally defined as a graph $\mathrm{G}=(V,E)$ with a object type mapping function $\phi: V \rightarrow \mathcal{A}$ and edge type mapping function $\psi: E \rightarrow\mathcal{R}$.
Here $v\in V$ are the nodes representing entities, $\phi(v) \in \mathcal{A}$ are the domains, $e \in E$ are the edges representing co-occurrence between entities and $\psi(e)\in \mathcal{R}$.
A \textbf{metapath}~\cite{sun2011pathsim} or metapath schema is an abstract path defined on the graph network schema of a heterogeneous network that describes a composite relation $\mathbf{R}=R_1 o R_2 \ldots R_l$ between nodes of type $A_1, A_2 \ldots A_{l+1}$, capturing relationships between entities of different domains.
There can exist multiple metapaths, and we consider $\mathcal{R}$ and thus the metapaths to be symmetric in our problem setting. 
Metapaths have been utilized in similarity search in complex data, and to find patterns through capturing relevant entity relationships~\cite{cao2018collective}.
Recent approaches on \textit{knowledge graph embeddings} (KGE)~\cite{wang2017knowledge} have demonstrated their effectiveness in capturing the semantic relationships between objects of different types in knowledge graphs which are HINs. 
Many approaches for KGE consider symmetric relationships as in our case, and is more generally applicable.
We choose one such model \textit{DistMult}~\cite{distmult2014} to obtain KGE for the entities in our data.
\textit{DistMult} uses both node and edge embeddings to predict semantically similar nodes, since it models relationships between entities in form of $<e_a, R , e_b>$.

In generating counterfactuals for an anomalous record, we intend to replace the entity $e_j$(or entities) which is predicted to have low likelihood by the explainer model, given the context comprising of the other entities in the record.
The intuition is to replace such entities with other entities (of the corresponding domain) which are semantically similar to the other entities in the record.
Let $\textbf{x}_a$ be the anomalous record and let entity $e_j^p$ in domain $d_j$ be selected for replacement. 
In this task, we utilize the associated HIN constructed from the data, along with the set of metapaths $MP = \{mp_1, mp_2 \ldots mp_q\}$ that are defined using domain knowledge.
Here metapath $mp_i$ is of the form $\{d_a, d_b \ldots d_i\}$.
Thus, candidates to replace $e_j^p$ are selected using the metapaths that contain $d_j$.
Let us consider one such metapath $mp_p$ such that $d_j \in mp_p$, with relations of $(d_i,d_j)$ and $(d_j,d_k)$. Let the respective entities in $\textbf{x}_a$ for $d_i$ and $d_k$ be $e_i^a$ and $e_k^a$.
In a generated counterfactual $\textbf{x}_{cf}$, the entity $e_j^{cf}$ that replaces $e_j^a$ should ideally be semantically similar to $e_i^a$ and $e_k^a$.
KGE can be effectively used for this task. This idea is described in Figure~\ref{fig:semantic}.
We find $K$ nearest entities to $e_i^a$ and $e_k^a$, belonging to domain $d_j$.
Note that it is possible that $d_i$ or $d_k$ is null based on the schema of $mp_p$.
We replace the entities in the domains with low likelihood with all combinations of the candidate replacements for the respective domains to obtain the set of candidate counterfactuals, of which $K$ least anomalous are chosen.
The steps are summarized in Algorithm~\ref{alg:recourse_gen_xformer}.

\section{Evaluation Metrics}\label{sec:metrics}
Evaluation metrics are crucial to understanding the performance of counterfactual generation methods, more so due to the fact that generated counterfactuals have multiple objectives and associated trade-offs. 
We discuss some of the metrics proposed in prior literature, and their limitations in the current problem setting.
Further, we propose a set of new metrics that are more appropriate.

\subsection{Existing Metrics for Counterfactuals}

\textbf{Recourse Correctness} or \textbf{validity}~\cite{mothilal2020explaining} captures the ratio of counterfactuals
that are accurate in terms obtaining the desired outcome from the blackbox prediction model.  
For unsupervised anomaly detection since $\mathcal{M}_{AD}$ provides a real valued likelihood (or anomaly score), a direct prediction (decision value) is unavailable. 
\textbf{Recourse Coverage}~\cite{rawal2020beyond} refers to quantification of the criterion that the algorithmic recourse provided covers as many instances as possible.
\textbf{Distance}~\cite{crupi2021counterfactual,karimi2020survey} or \textbf{proximity} measures the mean feature-wise distance between the original data instance and the set of recourse candidates. Distance is often calculated separately for categorical and continuous attributes. 
For continuous attributes $l_p$ norms~\cite{dhurandhar2018explanations} or their combinations are used whereas
for categorical(discrete) variables overlap measure~\cite{chandola2007similarity} $\mathbbm{1}(x_i=x_j)$ has been used. 
With purely categorical attributes, this measure however fails to convey any information other than merely how many of the attributes are different in the counterfactual.

\textbf{Cost}~\cite{crupi2021counterfactual,karimi2020model} refers to the cost incurred in changing a particular feature value in a recourse candidate. 
Prior works have utilized $l_p$ norms to quantify this criteria, for real-valued features.
In our problem scenario, this metric is directly not applicable without any external real-world constraints which can help quantify the difference in cost in changing $x_i$ to $x_j$ vs. $x_k$.
\textbf{Diversity}~\cite{mothilal2020explaining} refers to the feature-wise distances between the set of recourse candidates. 
Diversity encourages sufficient variation among the set of recourse candidates so that it increases the chance of finding a feasible solution. However, it has been noted that in certain cases diversity as an objective correlates poorly with user cost~\cite{yadav2021low}.
\textbf{Sparsity}~\cite{mothilal2020explaining} refers to the number of features that are different in the recourse candidates, with respect to the original data instance.

\subsection{Proposed Metrics}\label{sec:metrics_new}
We propose a new set of metrics based on previously defined metrics, which are more suited to our problem setting. 

\textbf{Sparsity-Index}: 
Sparsity is an important objective along with diversity that encourages minimal change is made to a data instance in terms of features.
To capture this notion, we define \textit{Sparsity Index} for tabular data with categorical features. Let $\textbf{x}_a$ be the anomalous record, and $d_j$ be the $j^{th}$ domain or feature.
\begin{equation}
    Sparsity \,Index = \frac{1}{\vert Y \vert}\Sigma_{\textbf{x} \in Y}\frac{1}{1 + \Sigma_{d_j}\mathbbm{1}(x_j \neq x_j^a) }
\end{equation}
The values of Sparsity Index $\in[0.5,1)$, with the low value corresponding to modification of all feature values and the maximum value corresponding to none.

\textbf{Coherence}: We define \textit{coherence} as measure to quantify the consistency of the counterfactuals similar to density consistency~\cite{karimi2020survey}.
Let $\mathcal{D}_p$ be the set of domains which are modified in $\textbf{x}_a$ to obtain a counterfactual $\textbf{x}_{cf}$, $\mathcal{D}_r$ be the remaining domains. Let $e_j$ be the entity in $\textbf{x}_{cf}$ for domain $j$.
\textit{Coherence} measures the mean probability of co-occurrence of the entities $e_i \in D_p$ with $e_j \in D_r$.
Maximizing coherence implies $e_j^r$ in $\textbf{x}_a$ is replaced with a candidate entity $e_j^{cf}$ in $\textbf{x}_{cf}$ which has a high probability of co-occurrence given the context of other entities of $D_r$ in $\textbf{x}_a$, and leads to plausible counterfactuals.
\begin{equation}
    coherence = \Sigma_{e_i \in D_p} \frac{1}{|\mathcal{D}_r|} \Sigma_{e_j \in \mathcal{D}_r}P(e_i,e_j)
\end{equation}

\textbf{Conditional Correctness}: This metric quantifies the validity of the counterfactuals, \textit{conditional} upon the underlying anomaly detection model $\mathcal{M}_{AD}$ which has a scoring function   $score_\mathcal{M}()$.
Let $\mathcal{D}_k$ be a set of randomly chosen data instances from the training and testing set. 
Let $\textbf{x}_a$ be the anomalous record and let the rank of $\textbf{x}_a$ in $\textbf{x}_a \cap \mathcal{D}_k$ be $r$, sorted by $score_\mathcal{M}()$ with appropriate order.
Without loss of generalization, we can assume a higher score indicates a more \textit{normal} or nominal data instance and a low score indicates anomalousness.
For $x \in Y$, where Y is the set of counterfactuals, \textit{conditional correctness} can be defined as 
\begin{equation}
    CC = \frac{1}{|Y|}\Sigma_{x \in Y} \mathbbm{1}(Rank(x) - r > 0 )
\end{equation}
This implies that $x$ is ranked lower in terms of being an anomaly, since higher ranked data instances are more anomalous.
Ranking is a more suitable approach to designing a metric than utilizing thresholds which are data and application dependant an is difficult to determine.
The relative ordering of records are important in this setting, since test instances are sorted based on $score_\mathcal{M}()$.

\textbf{Feature Accuracy}: The concept of anomaly in tabular data with categorical variables has been described as one or more attributes being \textit{out of context} with respect to the others, as discussed in Section~\ref{sec:preliminaries}. 
Therefore, it is important to accurately measure how well can a model identify which of the domain values should be modified in $\textbf{x}_a$, and relates to the explanation aspect of algorithmic recourse.
This requires having a Gold Standard (ground truth) knowledge where we know  which domain values (features) have been corrupted and the entities for those domains are out of context. 
Let $dom$ be the set of domains (features) with $m$ domains. 
Let $\mathbbm{q()}$ be a binary valued function that has value $1$ if the domain value is changed in counterfactual $\textbf{x}_{cf}$ from $\textbf{x}_a$ $(r_j \in \textbf{x}_a \rightarrow x_j\in \textbf{x}_{cf})$ is an actual cause of the anomaly or if a domain value remains unchanged if it was not a cause of the anomaly. 
\begin{equation}
    FA = \frac{1}{\vert Y \vert}\Sigma_{\textbf{x}_{cf}} \frac{1}{m} \Sigma_{j \in dom}\mathbbm{1}(\mathbbm{q}(r^j \rightarrow x^j )=1)
\end{equation}

\textbf{Heterogeneity}: Although diversity is an important objective for recourse candidates, existing diversity metrics like \textit{Count Diversity}~\cite{mothilal2020explaining} are inadequate.
A trivial random modification of all feature values will maximize such metrics for our setting with strictly categorical features, where distance between discrete feature values is computed using overlap measure. 
Two factors are important here: \begin{enumerate*}[label=(\roman*)]
    \item the variation among the entities that are proposed to replace original entity in $\textbf{x}_a$ and
    \item the correct domain's value is modified or not.
\end{enumerate*} 
We require the Gold Standard (ground truth) to determine whether the counterfactual modifies the a correct domain's value.
In our experiments, use of synthetic anomalies enables calculation of this metric.
Between any two pair of recourse candidates, \textit{heterogeneity} encourages dissimilarity while taking into account if both the pair of counterfactuals modify the correct domain's value. 
Let $m$ be the number of domains, $K$ be the size of the set of counterfactuals $Y$, and $\mathbbm{1}(\mathbbm{q})$ be 1 if the correct domain has been modified.
\begin{equation}
\centering
\begin{multlined}
    H = \frac{1}{K^2m}\Sigma_{i=1}^{K-1} \Sigma_{j=i+1}^{K} \Sigma_{l=1}^{m} w_{ij}^l \mathbbm{1}(x_i^l\neq x_j^l) \\
    w_{ij}^{l} = \mathbbm{1}(\mathbbm{q}(r^l\rightarrow x_i^l )=1) * \mathbbm{1}(\mathbbm{q}(r^l \rightarrow x_j^l )=1)
\end{multlined}
\end{equation}

\begin{table}[tp!]
    \centering
    \caption{Details of the datasets used for evaluation.}
    \begin{tabular}{c|c|ccc }
    \toprule
        Dataset & Source &\makecell{Total entity \\count} & \makecell{Domain \\Count} & \makecell{Train \\ size}\\
        \midrule
        Dataset-1 & US Import  &  6353 &  8 & 38291 \\
        Dataset-2 & US import  &  6151 &  8 & 35177 \\
        Dataset-3 & US import  &  7340 &  8 & 43495 \\
        Dataset-4 & Colombia Export &  4008 &  5 & 16758 \\
        Dataset-5 & Ecuador Export  &  3198 &  7 & 13956 \\
    \bottomrule
    \end{tabular}
    \label{tab:data}
\end{table}

\section{Empirical Evaluation}\label{sec:results}
The key objective here is to obtain counterfactuals for for anomalies in tabular data with categorical features.
We consider \textit{MEAD}~\cite{datta2020detecting} and \textit{APE}~\cite{chen2016entity} as the base anomaly detection models suited to categorical tabular data.
For our comparative evaluation against baselines we use \textit{MEAD}.
For an objective and quantifiable analysis of the performance of our approach with possible alternatives, we perform extensive experiments to capture the varied desiderata in terms of the metrics defined in Section~\ref{sec:metrics_new}. Further, we analyze the computational cost as well as the stability of the proposed approach.


\begin{table*}[htb!]
    \centering
    \caption{Comparison of performance of baselines and our approach for the adopted metrics.}
    \begin{subtable}[h]{0.99\textwidth}
     \centering
     \caption{Feature Accuracy}
      \begin{tabular}{p{2cm}||ccccc}
      \toprule
         Dataset    & Replace-m           & FIMAP              & RCEAA               & Xformer-R           & \textbf{CARAT} \\
        \midrule
         Dataset-1  & $ 0.8638 \pm 0.0858$ & $0.1897 \pm 0.0615$ & $0.2328 \pm 0.0414$ & $0.9803 \pm 0.0527$ & $0.9822 \pm 0.0509$\\
         Dataset-2  & $ 0.8581 \pm 0.0930$ & $0.1899 \pm 0.0617$ & $0.2585 \pm 0.0488$ & $0.9771 \pm 0.0641$ & $0.9813 \pm 0.0583$\\
         Dataset-3  & $ 0.8591 \pm 0.0876$ & $0.1898 \pm 0.0615$ & $0.2381 \pm 0.0580$ & $0.9769 \pm 0.0633$ & $0.9786 \pm 0.0630$\\
         Dataset-4  & $ 0.7167 \pm 0.1152$ & $0.2995 \pm 0.0962$ & $0.3094 \pm 0.0988$ & $0.9111 \pm 0.1388$ & $0.9164 \pm 0.1370$\\
         Dataset-5  & $ 0.5658 \pm 0.1566$ & $0.2181 \pm 0.0698$ & $0.2761 \pm 0.0542$ & $0.9577 \pm 0.0855$ & $0.9601 \pm 0.0831$\\
        \bottomrule
    \end{tabular}
    \label{tab:feature_accuracy}
    \end{subtable}
    \begin{subtable}[h]{0.99\textwidth}
      \centering
      \caption{Heterogenity}
      \begin{tabular}{p{2cm}||ccccc}
      \toprule
       Dataset   & Replace-m          & FIMAP              & RCEAA              & Xformer-R & \textbf{CARAT}   \\
        \midrule
         Dataset-1 & $0.3589 \pm 0.3056$ & $0.9737 \pm 0.0261$ & $0.6625 \pm 0.1098$ & $0.9693 \pm 0.0821$ & $0.9017 \pm 0.1244$\\
         Dataset-2 & $0.3728 \pm 0.3273$ & $0.9716 \pm 0.0278$ & $0.6646 \pm 0.1499$ & $0.9579 \pm 0.1311$ & $0.8930 \pm 0.1522$\\
         Dataset-3 & $0.3472 \pm 0.3164$ & $0.9737 \pm 0.0262$ & $0.6503 \pm 0.1236$ & $0.9662 \pm 0.1016$ & $0.8978 \pm 0.1319$\\
         Dataset-4 & $0.0634 \pm 0.1216$ & $0.9145 \pm 0.0965$ & $0.6863 \pm 0.2824$ & $0.8164 \pm 0.2333$ & $0.7805 \pm 0.2264$\\
         Dataset-5 & $0.1122 \pm 0.2041$ & $0.9570 \pm 0.0441$ & $0.6176 \pm 0.1957$ & $0.8879 \pm 0.1957$ & $0.8378 \pm 0.2003$\\
        \bottomrule
    \end{tabular}
    
    \label{tab:het}
    \end{subtable}
    \begin{subtable}[h]{0.99\textwidth}
      \centering
      \caption{Coherence}
      \begin{tabular}{p{2cm}||ccccc}
        \toprule
         Dataset   & Replace-m         & FIMAP             & RCEAA            & Xformer-R              & \textbf{CARAT}   \\
        \midrule
         Dataset-1 & $0.0012 \pm0.0005$ & $0.0000 \pm0.0000$ & $0.0002 \pm 0.0001$ & $0.0004 \pm 0.0003$ & $0.0025 \pm 0.0022$ \\
         Dataset-2 & $0.0008 \pm0.0004$ & $0.0000 \pm0.0000$ & $0.0002 \pm 0.0001$ & $0.0003 \pm 0.0002$ & $0.0020 \pm 0.0018$ \\
         Dataset-3 & $0.0013 \pm0.0006$ & $0.0000 \pm0.0000$ & $0.0002 \pm 0.0001$ & $0.0004 \pm 0.0003$ & $0.0030 \pm 0.0028$ \\
         Dataset-4 & $0.0003 \pm0.0004$ & $0.0000 \pm0.0000$ & $0.0001 \pm 0.0001$ & $0.0007 \pm 0.0012$ & $0.0018 \pm 0.0027$ \\
         Dataset-5 & $0.0007 \pm0.0007$ & $0.0000 \pm0.0001$ & $0.0005 \pm 0.0005$ & $0.0011 \pm 0.0014$ & $0.0037 \pm 0.0047$ \\
        \bottomrule
    \end{tabular}
    
    \label{tab:coherence} 
    \end{subtable}
    
    \begin{subtable}[h]{0.99\textwidth}
    \centering
    \caption{Sparsity-Index}
    \begin{tabular}{p{2cm}||ccccc}
    \toprule
     Dataset & Replace-m & FIMAP & RCEAA & Xformer-R & \textbf{CARAT}   \\
     \midrule
     Dataset-1 & $0.8889\pm0.0000$ & $0.5015\pm0.0010$ & $0.5355 \pm 0.0089$ & $0.8389 \pm0.0524$ & $0.8381 \pm0.0523$ \\
     Dataset-2 & $0.8889\pm0.0000$ & $0.5016\pm0.0010$ & $0.5391 \pm 0.0089$ & $0.8381 \pm0.0523$ & $0.8572 \pm0.0463$ \\
     Dataset-3 & $0.8889\pm0.0000$ & $0.5015\pm0.0010$ & $0.5321 \pm 0.0042$ & $0.8366 \pm0.0532$ & $0.8358 \pm0.0531$ \\
     Dataset-4 & $0.8750\pm0.0000$ & $0.5049\pm0.0022$ & $0.5445 \pm 0.0052$ & $0.7804 \pm0.0705$ & $0.7825 \pm0.0678$ \\
     Dataset-5 & $0.8333\pm0.0000$ & $0.5027\pm0.0014$ & $0.5468 \pm 0.0039$ & $0.8305 \pm0.0586$ & $0.8300 \pm0.0585$ \\
    \bottomrule
    \end{tabular}
    \label{tab:sparsity} 
    \end{subtable}
    
    \begin{subtable}[h]{0.99\textwidth}
    \centering
    \caption{Conditional Correctness}
    \begin{tabular}{p{2cm}||ccccc}
        \toprule
        Dataset    & Replace-m           & FIMAP              & RCEAA               & Xformer-R           & \textbf{CARAT}   \\
        \midrule
         Dataset-1 & $1.0000 \pm0.0000$  & $0.7404 \pm 0.1388$ & $0.6230 \pm 0.1609$ & $0.7389 \pm 0.1947$ & $0.9774 \pm 0.0631$\\
         Dataset-2 & $1.0000 \pm0.0000$  & $0.6646 \pm 0.1556$ & $0.4990 \pm 0.2092$ & $0.7414 \pm 0.2077$ & $0.9733 \pm 0.0800$\\
         Dataset-3 & $1.0000 \pm0.0000$  & $0.7544 \pm 0.1342$ & $0.6090 \pm 0.1489$ & $0.7634 \pm 0.2061$ & $0.9727 \pm 0.0679$\\
         Dataset-4 & $1.0000 \pm0.0000$  & $0.3666 \pm 0.2223$ & $0.3770 \pm 0.2712$ & $0.5572 \pm 0.2725$ & $0.8411 \pm 0.2429$\\
         Dataset-5 & $1.0000 \pm0.0000$  & $0.5122 \pm 0.1867$ & $0.4240 \pm 0.1744$ & $0.5858 \pm 0.2708$ & $0.8695 \pm 0.2023$\\
        \bottomrule
    \end{tabular}
    
    \label{tab:cc}
    \end{subtable}
    
    \begin{subtable}[h]{0.99\textwidth}
    \centering
    \caption{Summary (normalized mean) of all metrics across all datasets, for the baselines and our proposed approach \textit{CARAT}.}
    \begin{tabular}{cx{3cm} cx{3cm} cx{3cm} cx{3cm} cx{3cm}}
   
    \toprule
    Replace-m & FIMAP & RCEAA & Xformer-R & \textbf{CARAT}   \\
    \midrule 
    0.6150 & 0.2410 & 0.1657 & 0.6752& \textbf{0.9183} \\
    \bottomrule
    \end{tabular}
    \label{tab:summary}
    \end{subtable}
    \label{tab:metrics}
\end{table*}
\vspace{-0.25em}

    

   



    

   
    

\begin{table*}[hpt!]
    \centering
    \caption{Comparison of metrics for generated counterfactuals using different anomaly detection models with \textit{CARAT}.}
    \begin{tabular}{c||cc|cc|cc}
        \toprule
         & \multicolumn{2}{c}{Sparsity Index} & \multicolumn{2}{c}{Conditional Corr.} & \multicolumn{2}{c}{Coherence} \\
        \midrule
         Dataset & APE                              & MEAD                                & APE                             & MEAD                              & APE                             & MEAD   \\
        \midrule
         Dataset-1 & \makecell{$0.8764$  $\pm0.0362$} & \makecell{$0.8730$  $\pm0.0391$}  & \makecell{$0.7323$ $\pm0.3012$} & \makecell{$0.7237$ $\pm0.3283$} & \makecell{$0.0004$ $\pm 0.0004$} & \makecell{$0.0004$ $\pm0.0004$} \\
         Dataset-2 & \makecell{$0.8870$  $\pm0.0178$} & \makecell{$0.8695$  $\pm0.0404$}  & \makecell{$0.6624$ $\pm0.3003$} & \makecell{$0.6853$ $\pm0.3473$} & \makecell{$0.0003$ $\pm 0.0003$} & \makecell{$0.0004$ $\pm0.0004$} \\
         Dataset-3 & \makecell{$0.8810$  $\pm0.0370$} & \makecell{$0.8753$  $\pm0.0416$}  & \makecell{$0.7517$ $\pm0.2462$} & \makecell{$0.6326$ $\pm0.3721$} & \makecell{$0.0002$ $\pm 0.0002$} & \makecell{$0.0003$ $\pm0.0003$} \\
         Dataset-4 & \makecell{$0.8415$  $\pm0.0058$} & \makecell{$0.8194$  $\pm0.0517$}  & \makecell{$0.6422$ $\pm0.2703$} & \makecell{$0.5414$ $\pm0.3877$} & \makecell{$0.0004$ $\pm 0.0003$} & \makecell{$0.0003$ $\pm0.0004$} \\ 
         Dataset-5 & \makecell{$0.8720$  $\pm0.0238$} & \makecell{$0.8654$  $\pm0.0355$}  & \makecell{$0.7694$ $\pm0.2150$} & \makecell{$0.4601$ $\pm0.3657$} & \makecell{$0.0009$ $\pm 0.0008$} & \makecell{$0.0015$ $\pm0.0014$} \\
        \bottomrule
    \end{tabular}
   
    \label{tab:ad_model_comp}
\end{table*}
\subsection{Datasets}
The datasets used in for the experimental and evaluation setup are real world proprietary datasets of shipping records obtained from Panjiva Inc~\cite{panjiva-trade-data-2019}.
Specifically we use 5 datasets with no overlap, constructed from a larger corpus of records.
We consider records of a time period as the training set and subsequent time as the test set. 
We fix the test set size to 5000. 
The dataset details are described in Table~\ref{tab:data}.
The training set of each dataset is used to train the AD model as well as the KGE model.
Since we do not have ground truth data of anomalies, we use synthetic anomalies generated from the test set of each dataset following prior work~\cite{chen2016entity}, and allows us to analyze the results using the ground truth knowledge of what caused the record to be an anomaly.

\subsection{Competing Baseline Methods}\label{sec:baselines}
The area of algorithmic recourse specifically for anomaly detection is unexplored, and to the best of our knowledge only one prior work \textit{RCEAA}~\cite{RCEAA} exists on this. 
Prior work on algorithmic recourse deals with classification scenarios and they are not directly applicable to our setting. 
Moreover, as previously noted, most prior work deals with real valued or mixed valued data where the cardinality of categorical variables are significantly lower. 
Also, most prior approaches convert discrete variables to binary vectors through one-hot encoding and treat them as real valued vectors. 
We choose the following baselines for comparison:

    \textbf{Replace-m}: This approach generates an initial candidate set of all possible records by replacing the entity values in $m$ domains simultaneously, using all possible combinations. The records in this candidate set are scored by the given anomaly detection model, and $K$ top scored (least anomalous) records are considered as set of counterfactuals. We set $m=1$ due to computational limitations.  

    \textbf{FIMAP}~\cite{chapman2021fimap}: FIMAP is a model based approach for generating counterfactuals through adverserial perturbations using a \textit{perturbation network}, for a classification setting with known labels.
    To train the proxy classifier, a set of synthetic anomalies (assigned label $y=0$) and  normal instances (assigned label $y=1$).
    The \textit{perturbation network}, which generates counterfactuals, is trained by providing synthetic anomalies and passing the perturbed data instance to the pretrained proxy classifier, to obtain the desired label ($y=1$).
    
    \textbf{RCEAA}~\cite{RCEAA}: RCEAA uses an optimization based objective to exactly calculate a set of $K$ counterfactuals. Since the optimization requires real valued inputs, we adopt real-value relaxation on the one-hot encoded discrete representation of $\textbf{x}_a$ and use soft-threshold approach to obtain discrete outputs.

    \textbf{Xformer-R}: This method utilizes the explainer model to identify entities in an anomaly with low likelihood. Counterfactuals are generated by replacing the entity in the identified domains $\textbf{x}_a$ with entity values that are sampled uniformly from the domain.


\subsection{Results}
We present the results for the metrics discussed in Section~\ref{sec:metrics_new}. 
For conditional correctness, we sample a set of records containing both known synthetic anomalies and normal data instances. 
It is important to note that no single metric quantifies the different desiderata of the generated counterfactuals. 
Beginning with \text{feature accuracy}, which is reported in Table~\ref{tab:feature_accuracy} we see the approaches based on Transformer based explainer (\textit{Xformer-R} and \textit{CARAT}) have significantly better performance compared to the others.
This demonstrates that the explainer can effectively identify entities in records which do not conform to expected co-occurrence patterns.
For \textit{heterogeneity}, the results are presented in Table~\ref{tab:het}, while \textit{CARAT} performs well, but \textit{FIMAP} has somewhat better performance. 
This can be explained by the fact that counterfactuals generated by \textit{FIMAP} modify most of the entity values---which violates the sparsity objective.
Next we consider ~\textit{coherence}, which that captures how semantically similar the replaced entities are to the remaining ones in the counterfactuals generated.
As reported in Table~\ref{tab:coherence}, we find \textit{CARAT} performs significantly better. 
This implies the generated counterfactuals are consistent with the underlying data distribution.
Considering \textit{sparsity}, \textit{FIMAP} and \textit{RECEAA} have significantly lower values as shown in Table~\ref{tab:sparsity}, since the counterfactuals have multiple feature values modified from the given anomaly. \textit{Replace-m} has a high sparsity since a single entity is modified (m=1). \textit{Xformer-R} and \textit{CARAT} have similar performance in terms of sparsity.
Lastly, for \textit{conditional correctness} reported in Table~\ref{tab:cc} we find \textit{Replace-m} has perfect score due to performing exhaustive search for least anomalous records. CARAT shows competitive performance here, better than other baselines. 
Since no single metric comprehensively captures the requisite objectives that we are trying to maximize in generating counterfactuals---we summarize the model performances across all the metrics.
For each approach, we first obtain the average of the values across all datasets and then normalize them. 
Then we perform an unweighted average across all of these normalized metrics values to find a single performance value.
The results are reported in Table~\ref{tab:summary}, which shows \textit{CARAT} has a significant overall advantage. 

\begin{figure}[pt!]
    \centering
    \includegraphics[width=0.80\columnwidth]{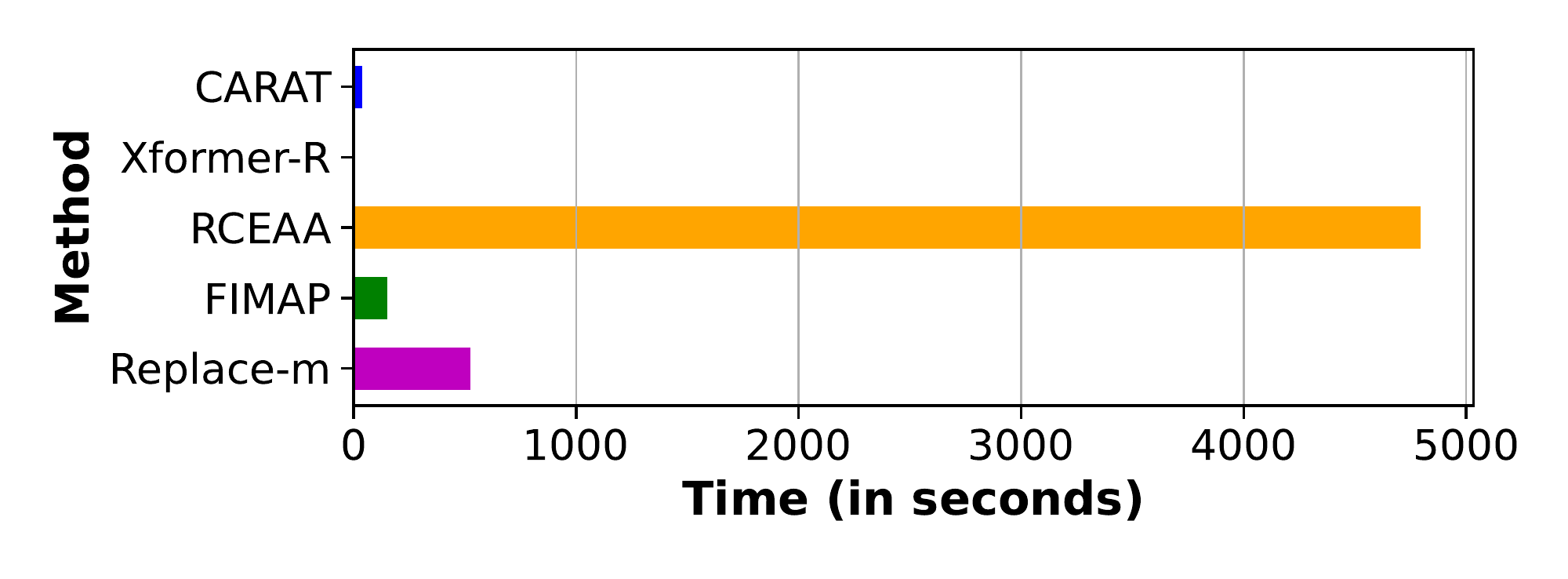}
    \caption{Computational time of compared approaches. Our proposed approach \textit{CARAT} shows a clear advantage.}
    \label{fig:time_complexity}
\end{figure}
\vspace{-0.5em}

\subsection{Stability}
The process of algorithmic recourse is inherently dependant on the underlying anomaly detection model which finds the anomalous data instances.
Thus it is important to understand the variation in performance of our proposed approach in the context of the underlying AD model $\mathcal{M}_{AD}$.
Here $\mathcal{M}_{AD}$ is a black-box model, and assumed to perfectly capture the underlying data distribution to find data instances that are true anomalies.
We choose two embedding based algorithms for tabular data with strictly categorical features as $\mathcal{M}_{AD}$, \textit{APE}~\cite{chen2016entity} and \textit{MEAD}~\cite{datta2020detecting}.
We apply \textit{APE} and \textit{MEAD} on test sets of each of the datasets, consider $1\%$ of the lowest scored records as anomalies, and apply \textit{CARAT} with the respective AD models on the corresponding anomalies.
Comparison of the applicable metrics defined in Section ~\ref{sec:metrics_new} for the generated counterfactuals across multiple metrics are reported in Table~\ref{tab:ad_model_comp}. 
We observe that similar performance is obtained in both cases, demonstrating the stability of our proposed approach.
\subsection{Computational Cost}\label{sec:time}
One of the major challenges of generating recourse is the computational complexity. 
We consider the computational cost in the execution phase, after all applicable pretraining and set up has been completed.
An approach with with exponential computational complexity would be simply infeasible in most practical scenarios. 
For instance the Replace-m approach outlined in Section~\ref{sec:baselines}, the complexity is $O(\Sigma_{1}^{l}\prod d_{i}, d_{i+1} .. d_{i+m})$, where $m$ domains are chosen to be modified at most, $d_{i}, d_{i+1} \dots d_{i+m}$ are the cardinalities of the $m$ domains, and $l = {|d| \choose m}$.
 \textit{RCEAA}~\cite{RCEAA} also suffers from high computational cost, as shown in Figure~\ref{fig:time_complexity}. 
The major bottlenecks in such an optimization based approach are:  
\begin{enumerate*}[label=(\roman*)]
\item expensive loss function
\item grid search for hyperparameters
\item operations on a very high dimensional vector due to one-hot encoding.
\end{enumerate*}
Our proposed approach is computationally efficient, since the explanation phase uses only a pretrained model with linear time complexity and the counterfactual generation phase that requires finding nearest neighbors is sped up using an indexing library~\cite{johnson2019billion}.

\begin{figure}[tp!]
    \centering
    \includegraphics[width=0.8750\columnwidth]{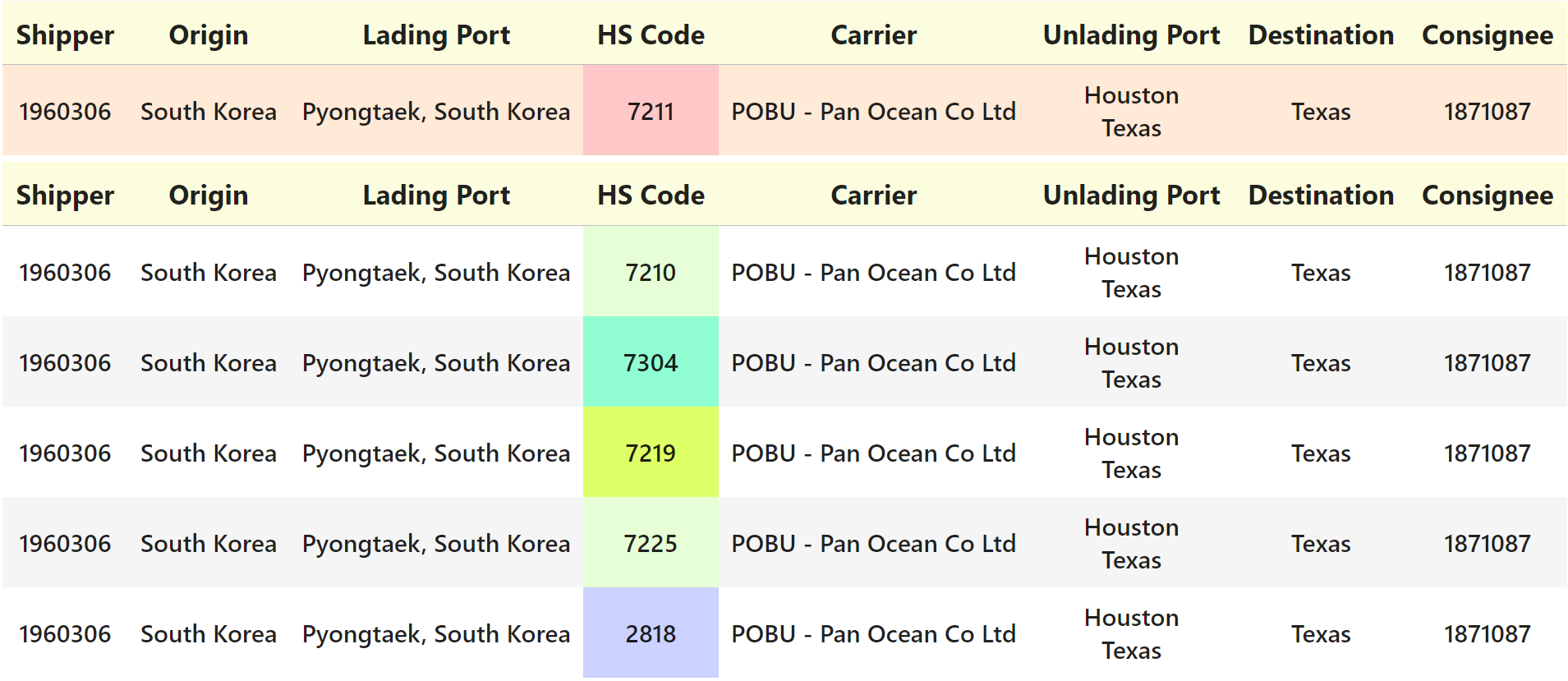}
    \caption{Example of an detected anomaly (first row) and the generated counterfactuals using \textit{CARAT}.}
    \label{fig:case_study}
\end{figure}
\vspace{-0.5em}

\subsection{Case Study of an Anomaly}
We perform a short case study based on one of the anomalies detected from test set of Dataset-1.
The detected anomaly and a set of counterfactuals are shown in Figure~\ref{fig:case_study}.
Our \textit{explainer} model finds that the entity in domain \textit{HS Code} in the anomalous record $\textbf{x}_a$ has low likelihood of occurrence in its context.
This is supported by the empirical data, as we find goods represented by the entity $<$HS Code:7211$>$ was previously traded by the neither the consignee (buyer) nor the shipper. Additionally, $<$HS Code:7211$>$ was not previously transported through the ports of lading and unlading in the record.
Our explainer only presents a low score for this entity, and not the others in the record although their context, which contains $<$HS Code:7211$>$, is altered. 
This demonstrates that the model can accurately predict likelihood from partially correct contextual information---which is the case for anomalous records.
Let us look at the counterfactuals, where $<$HS Code:7211$>$ is modified to alternate values.
$<$HS Code:7211$>$ refers to products of iron or non-alloy steel.
Firstly $<$HS Code:7210$>$, $<$HS Code:7225$>$ and $<$HS Code:7219$>$ refer to products very similar to $<$HS Code:7211$>$, specifically rolled or non-rolled steel or alloy products. 
$<$HS Code: 2818$>$ and $<$HS Code:7304$>$ are metal products as well. 
So the counterfactuals are essentially suggesting the buyer to obtain alternate products from the supplier (shipper). 
This can be explained based on the data, the particular type of goods $<$HS Code:7211$>$ are generally not sourced from the given origin and the shipper evidenced by any similar prior occurrences. 
A practical explanation might relate to industrial production and proficiency patterns, since certain regions specialize in specific products.
From prior records it is further observed that the consignee buys metal and alloy goods, such as with HS Codes 7210, 7323, 7304, 7225 and 7310.
Thus  \textit{CARAT} provides meaningful counterfactuals for detected anomalies.

\section{Conclusion and Future Work}
In this work we address the previously unexplored problem of algorithmic recourse for anomaly detection in tabular data with categorical features with high cardinality. 
We propose a novel deep learning based approach \textit{CARAT} to find counterfactuals for detected anomalies. 
We also define a set of relevant metrics that can enable effective evaluation of counterfactuals in such a problem setting.
The scalability and efficacy of our model in terms of the applicable metrics as well computational cost is demonstrated through extensive experiments.
However the current research leads to further research questions.
While we consider multiple objectives to optimize in the process of algorithmic recourse there are application scenario specific constraints that are not considered. 
One of the questions that has practical implications and requires domain knowledge is the actual cost incurred by the user. 
Another aspect is feasibility and actionability of counterfactuals, which are often dependant on extrinsic factors that need to explicitly considered and incorporated into the algorithmic recourse for anomalies.
Thus there are multiple continuing research directions which are a natural progression of the problem we address in this work.



\begin{acks}
This work was supported in part by US NSF grants CCF-1918770, NRT DGE-1545362, and OAC-1835660 to NR, and IIS-1954376 and IIS-1815696 to FC.
\end{acks}

\bibliographystyle{ACM-Reference-Format}
\bibliography{references}


\appendix
\section{Dataset Background}
The datasets used in the empirical evaluation are from shipping domain and are proprietary due to security and legal reasons.
We discuss some of the attributes of this real world data and their interpretations.

\textit{HS Code} or Harmonized Tariff Schedule Codes are globally standardized codes that define what type of goods are being transported. 
\textit{Carrier} is the transporting entity that operates between ports. 
The ports of \textit{lading} and \textit{unlading} are the points where the cargo is laden onto the transporting vessel or vehicle and and unladen from it.
We have received help of collaborating domain experts who deal with shipping data to help us understand the data characteristics and the relationships between attributes. The original data has many attributes which contain redundant information, and we select only meaningful attributes from the raw data. Also we remove rows with missing values and perform standard data cleaning to obtain our datasets.

The metapaths that describe these relationships are shown in Table~\ref{tab:metapaths}. These are designed with the knowledge of the structure of supply chains that are captured in this \textit{Bill of Lading} corpus.

\begin{table}[pb!]
    \centering
    \caption{Metapaths for the datasets belonging to the three data sources, viz. US Import, Colombia Export and Ecuador Export.}
    \begin{subtable}[h]{0.95\columnwidth}
    \centering
    \caption{Schema of the metapaths describing the relationships between attributes of the data for US Import }
    \begin{tabular}{l}
    \toprule
            Shipment Origin $\leftrightarrow$ HS Code $\leftrightarrow$Port Of Lading\\
            Shipment Destination $\leftrightarrow$ HS Code $\leftrightarrow$ Port Of Unlading\\
            Port Of Lading $\leftrightarrow$ HS Code $\leftrightarrow$ Carrier\\
            HS Code $\leftrightarrow$ Carrier $\leftrightarrow$Port Of Unlading\\
            Shipper $\leftrightarrow$ Shipment Origin $\leftrightarrow$ Port Of Lading\\
            Consignee $\leftrightarrow$ Shipment Destination$\leftrightarrow$Port Of Unlading\\
            Consignee $\leftrightarrow$ Carrier $\leftrightarrow$ Shipment Destination\\
            Shipper $\leftrightarrow$ Carrier $\leftrightarrow$ Shipment Origin\\
            Consignee $\leftrightarrow$ Carrier $\leftrightarrow$ Port Of Unlading\\
            Shipper $\leftrightarrow$ Carrier $\leftrightarrow$ Port Of Lading\\
     \bottomrule      
    \end{tabular}
    
    \label{tab:metapaths_US}
    
\end{subtable}

 \begin{subtable}[h]{0.95\columnwidth}
    \centering
    \caption{Schema of the metapaths describing the relationships between attributes of the data for exports from Colombia. }
    \begin{tabular}{l}
    \toprule
    Shipper $\leftrightarrow$ Shipment Origin $\leftrightarrow$ HS Code\\
    Consignee $\leftrightarrow$ Shipment Destination $\leftrightarrow$ HS Code\\
    Shipment Destination $\leftrightarrow$ HS Code $\leftrightarrow$ Shipment Origin\\
    Shipper $\leftrightarrow$ HS Code $\leftrightarrow$ Consignee \\
    \bottomrule      
    \end{tabular}
    
    \label{tab:metapaths_colombia}
\end{subtable}

 \begin{subtable}[h]{0.95\columnwidth}
    \centering
    \caption{Schema of the metapaths describing the relationships between attributes of the data for exports from Ecuador. }
    \begin{tabular}{l}
    \toprule
    Shipment Destination,Goods Shipped,Port Of Unlading\\
    Shipper $\leftrightarrow$ Goods Shipped $\leftrightarrow$ Shipment Origin\\
    Goods Shipped $\leftrightarrow$ Carrier $\leftrightarrow$ Port Of Unlading\\
    Consignee $\leftrightarrow$ Shipment Destination $\leftrightarrow$ Port Of Unlading\\
    Consignee $\leftrightarrow$ Carrier $\leftrightarrow$ Shipment Destination \\
    Shipper $\leftrightarrow$ Carrier $\leftrightarrow$ Shipment Origin \\
    Consignee $\leftrightarrow$ Carrier $\leftrightarrow$ Port Of Unlading \\
    \bottomrule      
    \end{tabular}
  
    \label{tab:metapaths_ecuador}
    \end{subtable}
    \label{tab:metapaths}
\end{table}
\section{Experimental Setup Details}

\subsection{Hardware and Libraries}
We provide the implementation details to faithfully reproduce the results obtained. 
All implementation is done in \textit{Python 3.9}, and uses standard libraries such as \textit{Numpy}, \textit{Pandas} and \textit{scikit-learn}.
For optimization and neural network based models, \textit{PyTorch} (version 1.10) is used.
All data preprocessing, training and evaluation presented in this work are performed on a 40-core machine, with a single GPU and distributed training required.
To train our Knowledge Graph Embedding model, we use the library \textit{StellarGraph}, which provides an implementation of \textit{DistMult}.

\subsection{Experimental Settings and Hyperparameters}
\subsubsection{Anomaly Detection Model}
Anomaly detection for tabular data with strictly categorical features, especially where the attributes have high dimensionality (cardinality) is a challenging task.
We choose Multi-relational Embedding based Anomaly Detection~\cite{datta2020detecting} as the base anomaly detection model for our experiments.
\textit{MEAD} uses an additive model based on shallow embedding, where the likelihood of a record is a function of the magnitude of transformed sum of the entity embeddings.
We use an embedding size of $32$ for anomaly detection models our experiments.

\subsubsection{Synthetic Anomalies}
Synthetic anomalies are generated using the approach followed in prior works~\cite{chen2016entity,datta2020detecting}. 
For each record, randomly one or more feature values are perturbed i.e. replaced with a random but valid feature value for the categorical attribute. Since our data has at most $8$ categorical attribute we limit the number of perturbations to 2. In generating counterfactuals we consider a balanced mix for all cases.

\subsubsection{CARAT Explainer Model Details}
For the \textit{explainer} in \textit{CARAT} presented in Section~\ref{sec:model} we an entity embedding dimension of 64. 
The encoder employs 4 layers of transformer blocks with 8 heads for multi-headed self-attention. 
The fully connected layers Decoder-R has 3 layers, with $256$,$128$ and $64$.
The fully connected layers Decoder-P has 3 layers, with $32$ and $16$.
We use the same architecture across all datasets.

In pretraining the encoder with decoder-R, the training objective is similar to Masked Language Model but not identical.
Specifically we replace approximately $20\%$ of entities in each records to \textit{mask}, and $20\%$ of entities are perturbed by replacement with a randomly sample entity from the same domain. 
In the second phase of training the \textit{decoder-P}, $\alpha$ --- the fraction of records which are not changed is set to $0.3$.

Both the pre-training phase of the encoder and the final explainer architecture are trained for 250 epochs with a batch size of 512 and learning rate of $0.0005$. 
All optimization for our model and the baselines are performed using Adam.

\subsubsection{CARAT KGE Details}
The knowledge graph embedding model adopted here is \textit{DistMult}. We use an embedding size of $100$.
The training batch size used is 1024, and we train the model for 300 epochs.
We use both the node and edge embeddings to find entities that are similar to a target entity. 
Since \textit{DistMult} uses $<$head,rel,tail$>$ format to calculate similarity, we perform nearest neighbor search for the tail entity using precomputed embedding of head nodes and relation type.

\subsection{Empirical Evaluation Setup}
We generate counterfactuals for synthetic anomalies. For each approach, we use a set of 400 anomalies and we generate 50 counterfactuals for each anomaly. 
We use the same set of anomalies for all approaches to perform a fair comparative evaluation.
For RCEAA we are able generate counterfactuals for 40 anomalies, due to the excessively long execution time as explained in Section~\ref{sec:time}.

\subsection{Additional Detail on Competing Baselines}
For the baseline models, we adapt the models to the current problem setting in an appropriate manner. 
We utilize the hyperparameters provided in the original work and do not perform significant hyperparameter tuning for these approaches. 
Similarly, we do not perform significant hyperparameter tuning for our model to tune performance as well since the objective is to demonstrate the validity of our approach in a general setting.

\subsubsection{RCEAA}
In our implementation of RCEAA~\cite{RCEAA}, we adapt the original approach.
Firstly, we replace the autoencoder based anomaly detection model with our likelihood based model.
In the loss function, we use $10^{th}$ percentile scores of training data as the threshold so that the generated counterfactuals have a low anomaly score (higher likelihood) according to our anomaly model. We set $\omega$ to 2. 
Additionally in place of euclidean distance, we use cosine distance since the dimensionality of the vectors is high.
In order to reduce the computational complexity due to grid search of hyperparameters, we set the upper and lower bounds of $\lambda_1$ and $\lambda_2$ to $0.25$ and $0.75$, and adopt step size of $0.25$. 
The training epochs are increased from $15$ mentioned in the paper, to $20$. We report the results with the lowest optimization loss.

\subsubsection{FIMAP}
For FIMAP, which is an approach for generating counterfactuals in a classification based setting, we adapt it to our problem setting appropriately. We adopt a more complex neural network architecture that follows that archetype proposed in the original work. Specifically, an embedding projection layer of dimensionality 32 is used for each categorical variable, whose output is concatenated and fed to a fully connected network for both proxy classifier network and perturbation network.
The fully connected network for proxy classifier has size $(256,256])$, following the original work that uses a 3 layered neural network.
The fully connected network for perturbation model has size $(256,128,128)$ and uses dropout of $0.2$. The value of $\tau$ in Gumbel softmax set to $0.5$.
We use $10000$ data points to train the proxy classifier for each dataset, with samples from training set and synthetic anomalies. The batch size used is 512, and the networks are trained for 100 epochs, with early stopping.

\section{Code and Data Link}
Please find the code for this work at : 

\url{https://github.com/ddatta-DAC/Algorithmic_Recourse_AnomalyDetection}

\section{Ethical Implications of Algorithmic Recourse for Anomaly Detection}
Since algorithmic recourse provides an approach towards generating counterfactuals that are not deemed anomalous by an anomaly detection model which may be part of a decision support system, it raises an obvious ethical question.
\textit{Is algorithmic recourse adverserial to anomaly detection i.e. intended to help nefarious actors attempting to evade detection?}

That is not our motivation here. First, data and anomaly detection models are expected to be secured and adverserial agents would have no access to them in order to circumvent the detection process.
Recourse for anomaly detection is intended to help the decision making process. It can enable organizations like enforcement agencies in our case study, that make use of anomaly detection systems to better handle false positive cases. Verified agents, whose transactions  may be erroneously flagged, could be intimated of the issue and may be provided alternatives or countermeasures. The issue of false positives exist since it is not feasible to always incorporate the application specific notions of anomaly into anomaly detection models. This effectively aids the decision support system user as well the agents whose data is being assessed.
In practical scenarios, systems that utilize algorithmic recourse do require transparency and human surveillance to ensure they are used as intended. There is a minimal risk, as in any system, that unscrupulous personnel who are insiders might utilize algorithmic recourse to provide alternatives to nefarious agents enabling them avoid detection. This however can be eliminated through correct operational and access protocols where such a system is deployed.



\end{document}